\documentclass[letterpaper, 10 pt, conference]{ieeeconf}  % Comment this line out
\IEEEoverridecommandlockouts                              % This command is only
\usepackage{graphicx}
\usepackage{amsmath,amssymb}
\usepackage{cite}
\usepackage{balance}
\usepackage{graphicx}
\usepackage{placeins}
\usepackage[ruled,vlined,linesnumbered]{algorithm2e}
\usepackage{algpseudocode}
\usepackage{enumerate}

\usepackage{multirow}
\usepackage{array}
\usepackage{booktabs}
\usepackage{threeparttable}
\usepackage{multirow}
\usepackage{gensymb}
\usepackage{subfig}
\usepackage{xcolor}

\usepackage{caption}
\usepackage{comment}

\newcolumntype{L}[1]{>{\centering\arraybackslash}m{#1}}
\renewcommand{\tabcolsep}{3pt}

\usepackage{caption}
\title{\LARGE \bf Long-Horizon Motion Planning via Sampling and Segmented Trajectory Optimization}

\author{ Jessica Leu, Michael Wang, and Masayoshi Tomizuka% <-this % stops a space
    \thanks{All authors are with the Department of Mechanical Engineering, University of California,
    Berkeley, CA 94720 USA {\tt\small jess.leu24, mwang542, tomizuka@berkeley.edu}
    }
}

\begin{document}

\maketitle
\thispagestyle{empty}
\pagestyle{empty}

%%%%%%%%%%%%%%%%%%%%%%%%%%%%%%%%%%%%%%%%%%%%%%%%%%%%%%%%%%%%%%%%%%%%%%%%%%%%%%%%
\begin{abstract}
This paper presents a hybrid robot motion planner that generates long-horizon motion plans for robot navigation in environments with obstacles. We propose a hybrid planner, RRT* with segmented trajectory optimization (RRT*-sOpt), which combines the merits of sampling-based planning, optimization-based planning, and trajectory splitting to quickly plan for a collision-free and dynamically-feasible motion plan. When generating a plan, the RRT* layer quickly samples a semi-optimal path and sets it as an initial reference path. Then, the sOpt layer splits the reference path and performs optimization on each segment. It then splits the new trajectory again and repeats the process until the whole trajectory converges. We also propose to reduce the number of segments before convergence with the aim of further reducing computation time. Simulation results show that RRT*-sOpt benefits from the hybrid structure with trajectory splitting and performs robustly in various robot platforms and scenarios.  
\end{abstract}

%%%%%%%%%%%%%%%%%%%%%%%%%%%%%%%%%%%%%%%%%%%%%%%%%%%%%%%%%%%%%%%%%%%%%%%%%%%%%%%%
\section{INTRODUCTION}
\begin{figure}[t]
\begin{center}
\includegraphics[width=8cm]{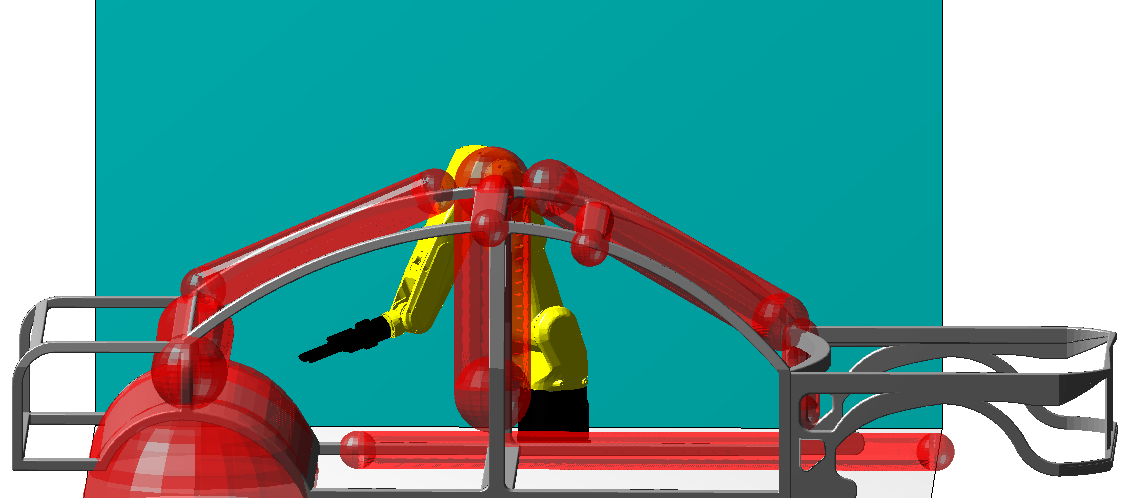}
\caption{A manipulator navigating through a car frame in a factory.}
\label{fig:car}
\end{center}
\end{figure}

Motion planning is one of the key challenges in robotics  \cite{lavalle2006planning}. It refers to the problem of finding a collision-free and dynamically-feasible path between the initial configuration and the goal configuration in environments full of obstacles (Fig.~\ref{fig:car}). Existing motion planning algorithms fall into two categories: planning-by-construction or planning-by-modification \cite{Acsf2017}. Search-based planning and sampling-based planning are two typical plan-by-construction algorithms. Algorithms such as A* and D* search \cite{Astar1968, Stentz94optimaland} belong to search-based algorithms, whereas rapidly-exploring random tree (RRT) \cite{Lavalle98rapidly}, probabilistic roadmap (PRM) \cite{prm1996}, and their variations \cite{webb2013kinodynamic,karaman2011sampling,birrt2000, islam2012rrt,bohlin2000path,amato1998obprm} belong to sampling-based planning. Planning-by-modification refers to algorithms that reshape a reference trajectory to obtain optimality regarding specific properties. Optimization-based algorithms belong to this category  \cite{Spellucci1998ANT,liu2018convex, chomp2009, trajopt2014, leu2020safe}.

Among the different planning problems, long-horizon motion planning is especially challenging in terms of finding a feasible solution and improving the quality of the solution. With only a naive initialization (e.g., a straight line from the initial to the goal in the configuration space), optimization-based planners struggle to find a solution that can travel long distances and make multiple big turns, which are often needed in long-horizon motion planning \cite{likhachev2009planning}. On the other hand, search-based methods require more memory space to store the precomputed graph; and a well designed heuristic must be provided to guide the search. In this light, search-based methods face larger challenges when the robot has many degrees of freedom and when generating an effective heuristic is difficult due to the cluttered environment \cite{dailong2021}. Also, the precomputed graph and the heuristics are hard to reuse if the obstacle configuration in the environment is changed.

A strong candidate for long-horizon path planning is RRT*, which is known for it's efficient, exploratory property of finding a feasible and semi-optimal \cite{karaman2011sampling, islam2012rrt} path with the shortest time. Although RRT* can converge to the optimal solution given infinite computation time, users often terminate the algorithm at a time limit, resulting in an unsmooth path. In addition, the constructed path often does not consider the robot dynamics, since deliberately considering these constraints weakens the computation advantage. Most RRT* variations proposed lately \cite{sandakalum2022motion} remain to be ``path planners'' rather than ``motion planners.'' In contrast, planning-by-modification methods are known to be very efficient given a good initialization. This motivates the development of hybrid motion planners \cite{Acsf2017,birrtopt2016,RMTrajopt2018,RMTrajopt2015,leu2021efficient}, which use planning-by-construction methods to generate a feasible reference path and use optimization-based methods to polish the solution.     
    
One of the computational bottlenecks of hybrid motion planners is the optimization step. An optimization-based motion planner that does not exploit the structure of the planning problem would normally scale in complexity with cube or square of the planning horizon $N$, i.e., $\mathcal{O}(N^3) \sim \mathcal{O}(N^2)$ \cite{o2013splitting, nielsen2015parallel}. This is especially costly when solving long-horizon planning problems where the number $N$ is large. To mitigate this problem, researchers propose splitting the problem into several sub-problems and develop an iterative update strategy to combine the distributed solutions and find the optimal trajectory \cite {sindhwani2017,singh2019inducing,wang2021trajectory}. This formulation also allows the solver to exploit parallel computation power. 

In this work, we incorporate the hybrid planner structure and propose RRT*-sOpt with the focus on long-horizon motion planning scenarios. We develop a segmented trajectory optimization (sOpt) layer that segments the initial reference from the RRT* layer equally by time and performs optimization iteratively and efficiently. We also notice that the optimal number of segments can vary as the optimization process goes. In other words, ``merging'' segments can be beneficial to the computation time because it enforces consensus between neighboring segments and may reduce the number of iterations needed before the overall trajectory converges. Therefore, we identify the conditions for merging segments and demonstrate its effectiveness.    
%CHOMP and TrajOpt which use sequential quadratic programming (SQP) and gradient descent, respectively—are at least $\mathcal{O}(N^2)$ in computational complexity with regard to the number of waypoints. In long-horizon motion planning especially, computation times can be lengthy. In order to still capitalize on the performance of optimization-based methods, parallel or distributed optimization techniques can be applied. In this work, we propose RRT*-sOpt, to solve long-horizon motion planning problem efficiently. 

Compared to our previous work, RRT*-CFS \cite{leu2021efficient}, we improve the optimization layer to exploit the parallel computation power. %This proposed method of “trajectory splitting” exploits the observation that time complexity decreases as path length decreases. 
Simulation results show that RRT*-sOpt provides a significant improvement in computation time, along being more robust at successfully finding long-horizon motion plans in complex environments.
Our main contributions are threefold as follows:
\begin{itemize}
  \item We develop a segmented trajectory optimization strategy (sOpt) with a segment merging scheme.
  \item The proposed RRT*-sOpt algorithm improves upon RRT*-CFS in runtime and robustness.
  \item We implement RRT*-sOpt and demonstrate its success with extensive simulation on multiple robot platforms.  
\end{itemize}

The remainder of the paper is organized as follows. Section 2  discusses the related works. Section 3 presents our proposed algorithm. Section 4 presents the simulation results  (video is publicly available at {\tt\small jessicaleu24.github.io/ECC2022.html}). Finally, we conclude the work in Section 5.

%%%%%%%%%%%%%%%%%%%%%%%
\section{PROBLEM FORMULATION and RELATED WORKS}\label{Related_works}
\subsection{Baseline Problem Formulation}
\label{OP}
In many scenarios, robot motion planning can be performed by solving an optimization problem with the following form:
\begin{equation}\label{eq:mainP}
\min_{\mathbf{x} \in \Gamma} f(\mathbf{x}),	
\end{equation}
where $\mathbf{x} \in \mathbb{R}^n$, and $\Gamma$ defines the feasible set:
\begin{equation}
\label{eq:cons}
	\Gamma = \bigcap_{j}\Gamma_{j} = \bigcap_{j}\{\mathbf{x}: h_{j}(\mathbf{x}) \ge 0\}.
\end{equation}
We assume that the constraint function $h_{j}(x)$ is a semi-convex function \cite{liu2018convex}. For example, $h_j(x)$ can be a safety function that measures the distance between a robot and the $j$th obstacle. The cost function, $f: \mathbb{R}^n \rightarrow \mathbb{R}$, is strongly convex and smooth. Note that the motion planning problem is non-convex due to the existence of the obstacles and the non-linear robot kinematics. Also, the dimension of the planning problem, $n$, depends on both the robot's state space and the number of waypoints from the initial state to the final state; $n$ is especially large in long-horizon motion planning problems. Therefore, solving motion planning problems is hard in general.

\subsection{Hybrid Motion Planning Algorithms}
Many works have focused on hybrid planners \cite{Acsf2017, birrtopt2016, RMTrajopt2018, RMTrajopt2015,leu2021efficient}. Methods such as lattice A* search, bidirectional RRT \cite{birrt2000}, or roadmaps are commonly used in the planning-by-construction stage; while methods such as SQP \cite{Spellucci1998ANT}, CFS \cite{liu2018convex}, and TrajOpt \cite{trajopt2014} are often used to polish the solution. Hybrid planners have better computation time efficiency than non-hybrid ones and can also solve harder problems, such as the narrow passage problem \cite{leu2021efficient}. We previously proposed a hybrid planner, RRT*-CFS, and demonstrated its computational speed advantages over its counterparts. Although its performance is robust with the test cases in the paper, it wanes when applied to long trajectories and higher-dimensional problems. This is also a common problem for most hybrid planners because the problem complexity normally scales with cube or square of the planning horizon and the robot state space. Therefore, we develop RRT*-sOpt to mitigate this problem. 

 %We propose an improvement to the optimization layer: the path-planning problem is segmented by time into subproblems of equal density, and an update scheme is applied until the solution converges to near-optimum. By distributing complexity, this proposed method of “trajectory splitting” exploits the observation that complexity decreases as path length decreases. We observe that RRT*-sCFS provides a significant improvement in computation time, along being more robust at successfully finding trajectories in complex environments.

\subsection{Segmented Trajectory Optimization Algorithms}
 In recent years, researchers have proposed to develop planners that enable the exploitation of parallel computing with multi-core CPUs/GPUs. Many have utilized alternating direction method of multipliers (ADMM) \cite{admm} to solve a highly non-linear and non-convex problem in a distributed manner. \cite {sindhwani2017} proposed to split the problem into two subproblems that consider dynamic and collision avoidance constraints, respectively, and combine the two solutions with a consensus update. Nevertheless, little speed-up was gained since the amount of waypoints (i.e., time steps) in the subproblems remained the same. In addition, this splitting method may not split the complexity evenly, which results in wait-time for the more complex process to finish \cite{wang2021trajectory}. \cite{singh2019inducing} achieved a distributed structure by decomposing the mobile-manipulator trajectory optimization into a sequence of convex QPs. However, collision avoidance was not demonstrated in this work. \cite{wang2021trajectory} proposed a similar distributed formulation for robot motion planning with collision avoidance. Yet, all of these works do not consider scenarios that require long-horizon motion planning and may still suffer with naive initialization. In this work, we leverage the hybrid structure to obtain an initial reference and focus on exploiting parallel computation power using sOpt for long-horizon motion planning with collision avoidance.
 %While convergence and robustness to poor initialization were empirically validated with their simulations, a more formal understanding of the convergence analysis was not presented.

%For the case of a mobile manipulator, a highly non-convex trajectory optimization problem, the authors of \cite{singh2019inducing} employ the use of an ADMM \cite{admm} based approach and other reformulations to achieve distributiveness. First, this augmented Lagrangian reformulation adds an additional constraint penalty with a Lagrange multiplier onto the original Lagrangian. Next, a convex surrogate is derived for each of the non-convex steps, which notably, can be split into multiple parallel optimizations based on the length of the time horizon. The convergence of their trajectory optimization was empirically validated by observing the maximum consensus residual at each iteration. Robustness to poor initial guess was also empirically validated. However, a more formal understanding of the convergence analysis for their proposed convex surrogate was not presented.

\section{THE PROPOSED ALGORITHM}
\begin{figure*}
 \centering        
      \begin{tabular}{@{}cc@{}}
  
   \begin{minipage}{.31\textwidth}
    \includegraphics[width=\textwidth]{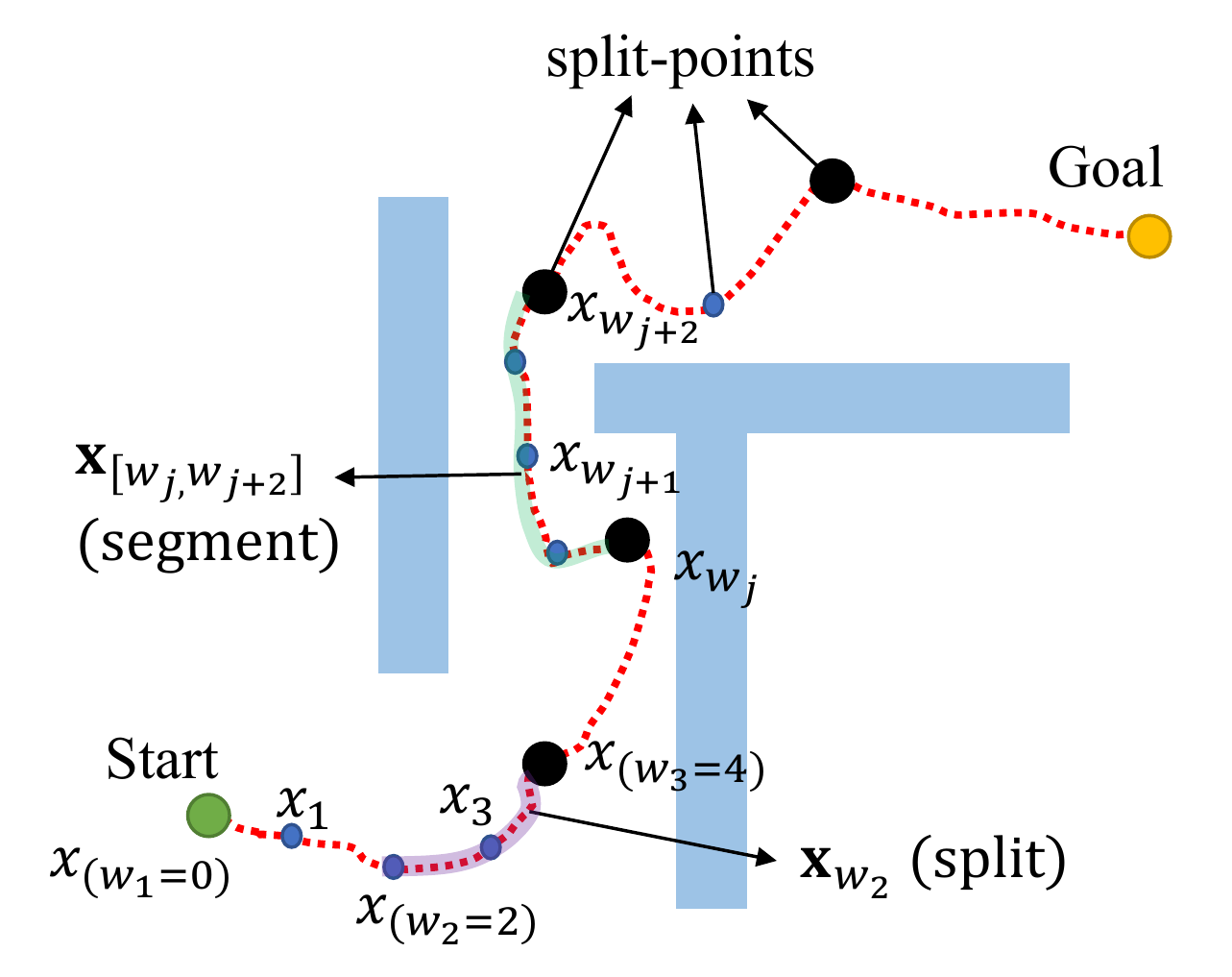}
   
   \end{minipage} &
    \begin{minipage}{.67\textwidth}
    \includegraphics[width=11.8cm]{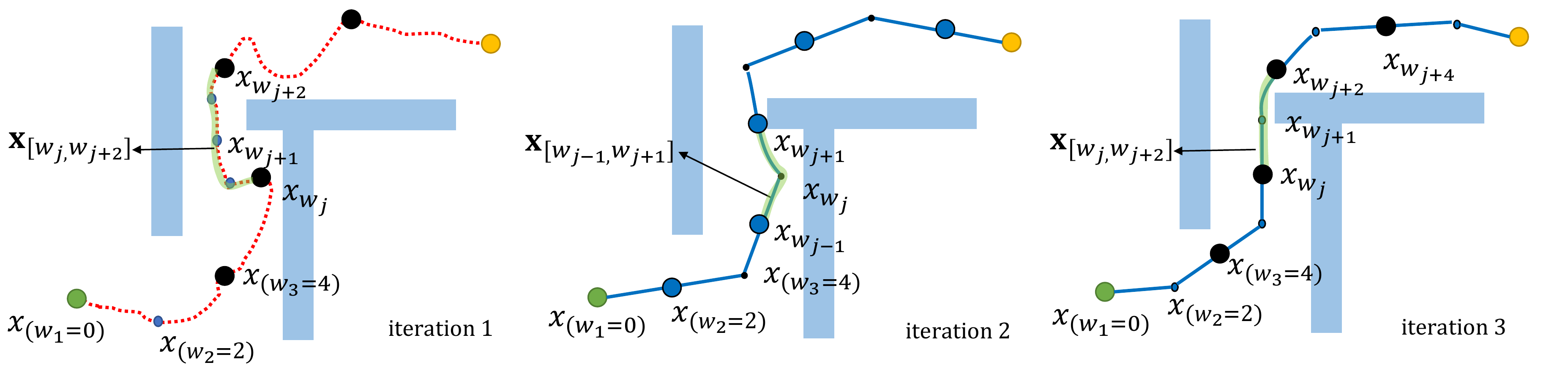}
   
   \end{minipage} 
      
  \end{tabular}
        \caption{The illustration of a segmented trajectory (left most) and illustrations of the segment selection for the first three iterations during sOpt (the other tree figures on the right).}
  \label{fig:splitting}
\end{figure*}

The proposed RRT*-sOpt inherits the merits of hybrid planners and segmented trajectory optimization. The RRT*-sOpt algorithm solves the non-convex motion planning problem by first quickly finding a feasible and semi-optimal path, and then iteratively refining the solution using sOpt. The RRT*-sOpt has three main features.
\begin{itemize}
  \item RRT*-sOpt has stochasticity due to the random sampling process in RRT*. This helps RRT*-sOpt to find a feasible path and avoid bad local optima that optimization-based algorithms may suffer from.
  \item Both the RRT* layer and the sOpt layer can be implemented with parallel computation. This allows us to significantly reduce the computation time.
  \item sOpt leverages parallel computation to mitigate the high dimensionality of long-horizon planning problems and implements a segment merging strategy to further reduce the computation time.  
\end{itemize}   
The RRT*-sOpt algorithm is summarized in Algorithm~\ref{alg:RRTsOpt}. We introduce the details of the proposed method in the following sections.

\begin{algorithm} \small
    \bf{input} {$\theta_{0}, \theta_{goal}, n_{samples}, \mathcal{O}, N_{segments}$}\\
    \While{! $\exists$ $\pmb{\theta}^{RRT^*}$}{
	$\pmb{\theta}^{RRT^*} \gets \tt{parallel\_RRT^*}$$(\theta_{0}, \theta_{goal}, n_{samples}, \mathcal{O})$\\
	}
    $\mathbf{x}^{(0)} \gets 
    \tt{generate\_reference}$$(\pmb{\theta}^{RRT^*})$ \\
    $\mathbf{W}^{(0)}\gets \tt{split\_reference}$$(\mathbf{x}^{(0)}, N_{segments})$\\
	\While {termination conditions not met}{
	    $mode \gets \tt{odd\_or\_even}$$()$\\
	    \For{$w^{(k)}_j \in \mathbf{W}^{(k),mode}$}{
	        $\chi_{[w_{j}, w_{j+2}]}^{(k)} \gets \tt{Obs\_select}$$(\mathbf{x}_{[w_{j}, w_{j+2}]}^{(k)}, \mathcal{O})$\\
	        $\mathbf{x}_{[w_{j}, w_{j+2}]}^{(k+1)} \gets \tt{OPT}$$(\mathbf{x}_{[w_{j}, w_{j+2}]}^{(k)},\chi_{[w_{j}, w_{j+2}]}^{(k)})$\\
	        % Should I include obstacle selection here?
	    }
	    \For{$w_{j}^{(k+1)} \in \mathbf{W}^{(k),mode}$} {
	    $d \gets \tt{iter\_prog}$$(\mathbf{x}_{[w_{j}, w_{j+4}]}^{(k)},\mathbf{x}_{[w_{j}, w_{j+4}]}^{(k+1)})$\\
	        \If{$d \leq \frac{2\epsilon H }{N_{seg}}$}{Remove $w^{(k)}_{j+2}$ from $\mathbf{W}^{(k)}$}
	    }
	    $\mathbf{x}^{(k+1)} \gets \tt{resample\_traj}$$(\mathbf{x}^{(k+1)})$\\
	    $\mathbf{W}^{(k+1)} \gets \tt{split\_reference}$$(\mathbf{x}^{(k+1)},\mathbf{W}^{(k)})$\\
	    $k \gets k+1$\\
	}
    \Return $\mathbf{x}^{(k)}$
	\caption{RRT*-sOpt}
	\label{alg:RRTsOpt}
\end{algorithm}

\subsection{The parallel RRT*}
Denote the configuration of a $d$-degree-of-freedom ($d$-DoF) robot as $\theta \in \mathbb{R}^d$, the initial configuration as $\theta_0$, the goal configuration as $\theta_{goal}$, the maximum number of samples in one RRT* thread as $n_{samples}$, the obstacles as $\mathcal{O}$, and the initial number of segments as $N_{segments}$. In Algorithm~\ref{alg:RRTsOpt}, the planner first runs parallel RRT* until it finds a feasible path that connects the initial configuration and the goal configuration. If more than one thread find a path, we choose the shortest path and set it as $\pmb{\theta}^{RRT^*}$. By setting up the $n_{samples}$ properly, we can find a solution in the first batch almost every time. To provide an initialization for the optimization layer, we first calculate path length of $\pmb{\theta}^{RRT^*}$, and then find the appropriate planning horizon $H$ according to the desired robot operation speed. Let the robot states be $x\in \mathbb{R}^n$; we generate the initial reference $\mathbf{x}^{(0)}:= [x_0^{\top}, x_1^{\top} ,\dots,x_{H}^{\top}]^\top$ using the sampled path from $\pmb{\theta}^{RRT^*}$. This process can be done by feeding the $\pmb{\theta}^{RRT^*}$ to a motion generator (e.g., iLQR \cite{tassa2012synthesis}) that outputs a motion plan, $\mathbf{x}^{(0)}$, which is a trajectory that follows the RRT* path.

\subsection{The segmented trajectory optimization}

The second part of the algorithm is the sOpt layer, which solves the planning subproblems iteratively. 

\noindent\textbf{Trajectory segmentation.} An illustration of the terminology for sOpt is shown at the left-most figure in Fig.~\ref{fig:splitting}. We denote a split of the trajectory as $\mathbf{x}_j$ (the purple line in Fig.~\ref{fig:splitting}). Given an integer number $N$, a trajectory with $2N$ splits is $\mathbf{x}:= [ \mathbf{x}_1,\dots,\mathbf{x}_j,\dots,\mathbf{x}_{2N}]$, each containing $(H-1)/2N+1$ waypoints. The indices of the split-points, i.e., the indices of the starting and ending point of each split are stored in a set $\mathbf{W} = \{ w_1,w_2,\dots, w_j,\dots,w_{2N+1} \}$ where $w_j \in \{0,1,\dots,H\}$. A segment (the green line in Fig.~\ref{fig:splitting}) contains two splits, denoted as $\mathbf{x}_{[i,i+(H-1)/N]}$. To ensure connectivity between segments, the end point of the previous segment is set to be the same as the first point of the succeeding segment. As shown in the right part of Fig.~\ref{fig:splitting}, in the odd iterations, i.e. $k$th iteration where $k=1,3,\dots$, all the segments start with index in the odd entries of $\mathbf{W}$, i.e., $\mathbf{W}^{odd}$. In the even iterations, segments start with index in the even entries of $\mathbf{W}$, i.e., $\mathbf{W}^{even}$.  For example, if $N = 3$, the first set of segments are $\{\mathbf{x}_{[w_1,w_3]}, \mathbf{x}_{[w_3,w_5]}, \mathbf{x}_{[w_5,w_7]}\}$; then, it becomes $\{\mathbf{x}_{[w_2,w_4]}, \mathbf{x}_{[w_4,w_6]}, \mathbf{x}_{w_1}, \mathbf{x}_{w_6}\}$ in the next iteration. The last two elements are the splits at the beginning and the tail.

\begin{algorithm}
\caption{Obstacle selection}
\label{alg:Obs}
\bf{input} {$\mathbf{x}_{[w_{j}, w_{j+2}]}^{(k)},\mathcal{O}$}\\
        $\tt{Obs\_getID(\cdot)} \gets \tt{Watershed}(\mathcal{O})$\\
        \For {$x_i \in \mathbf{x}_{[w_{j}, w_{j+2}]}^{(k)}$} {$id_i \gets \tt{Obs\_getID}$$(x_i)$}
        $\chi_{[w_{j}, w_{j+2}]}^{(k)} =\bigcap_{id}\{ \mathbf{x}: h_{id}( \mathbf{x}_j^{(k)}) + \nabla^{\top} h_{id}( \mathbf{x}_j^{(k)})( \mathbf{x} -  \mathbf{x}_j^{(k)}) \ge 0\}$\\
        \Return $\chi_{[w_{j}, w_{j+2}]}^{(k)}$
\end{algorithm}

\noindent\textbf{Segmented trajectory optimization.} The obstacle avoidance constraints of the planning problem are also distributed to each segment. As shown in Algorithm~\ref{alg:Obs}, in the $k$th iteration, we utilize the function $\tt{watershed}$ from MATLAB to select nearby obstacles for each segment. The safety functions associated with these obstacles are linearized at the reference segment, $\mathbf{x}_{[w_{j}, w_{j+2}]}^{(k)}$, to formulate the constraints as a convex feasible set, $\chi_{[w_{j}, w_{j+2}]}^{(k)}$ \cite{liu2018convex}. The planning subproblem optimizes each segment according to a cost function that has the form: $f(\cdot) = \| \mathbf{x}-\mathbf{x}_{goal}\|_{Q}^2 + \lambda  \|\mathbf{x}\|_R^2$. By fixing the initial and the goal waypoints to stay at $x_{w_{j}}$ and $x_{w_{j+2}}$ respectively, we formulate the subproblem as follows: 
\begin{equation}\label{P:MP}
\begin{aligned}
\mathbf{x}_{[w_{j},w_{j+2}]}^{(k+1)} = \arg\min_{\mathbf{x}}\quad  &f(\mathbf{x}),\\
s.t.\quad & f_{ki}(\mathbf{x}) \in \chi_{[w_{j}, w_{j+2}]}^{(k)},\\
\quad & x_0 = x_{w_{j}},\\
\quad & x_{(H-1)/N} = x_{w_{j+2}},
\end{aligned}
\end{equation}
where $f_{ki}(\cdot)$ is the robot kinematic model. Note that the selection of the starting and ending points alternates in each iteration so that the waypoints fixed in the present iteration will be optimized in the next iteration, and ultimately, the full trajectory can be optimized iteratively. 

\begin{figure}[t]
\begin{center}
\includegraphics[width=.9\linewidth]{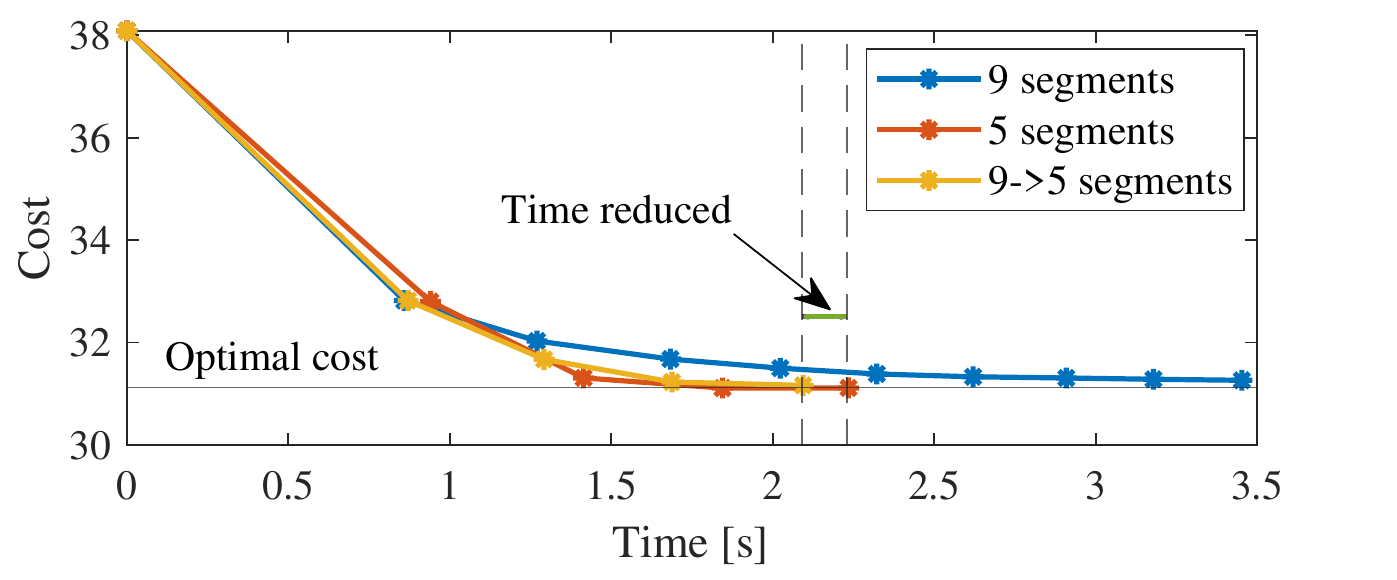}
\caption{An example of computation time reduced by segment merging in 2D planning. (Notice that we hard coded this merge to generate this plot for clearer visualization. In the simulations, merging normally occurs in later iterations.)}
\label{fig:merging_data}
\end{center}
\end{figure}

\noindent\textbf{Merging segments.} We observe that performances of sOpt with different numbers of segments are different at different stages. As shown in Fig.~\ref{fig:merging_data}, a sOpt with more segments reduces the cost quickly at the beginning but converges slowly later on (blue line) compared to a sOpt with fewer segments, which performs in the opposite way (orange line). Therefore, we propose to merge neighboring segments in later iterations when quick convergence is desired. As shown in Algorithm~\ref{alg:RRTsOpt} line $11\sim14$, the function $\tt{iter\_prog}$ calculates the cost of two neighboring segments and compares it with the cost of the previous iteration to quantify the progress made by the optimization in that iteration, i.e., 
\begin{equation}
\label{eq:merge}
	iter\_prog=\|f(\mathbf{x}_{[w_{j}, w_{j+4}]}^{(k)})-f(\mathbf{x}_{[w_{j}, w_{j+4}]}^{(k+1)})\|.
\end{equation}
By selecting a threshold $\epsilon$, merging happens when $\tt{iter\_prog}$ $ \leq 2\epsilon H/ N$.

%a function:
%\begin{equation}
%\label{eq:merge}
%	iter\_prog(\cdot)=\sum_{i\in ({w_j}^{(k)}, w_{j+4}^{(k)})} %\|x^{(k)}_i - x^{(k+1)}_i\|_2,
%\end{equation}

\begin{figure}[t]
\begin{center}
\includegraphics[width=\linewidth]{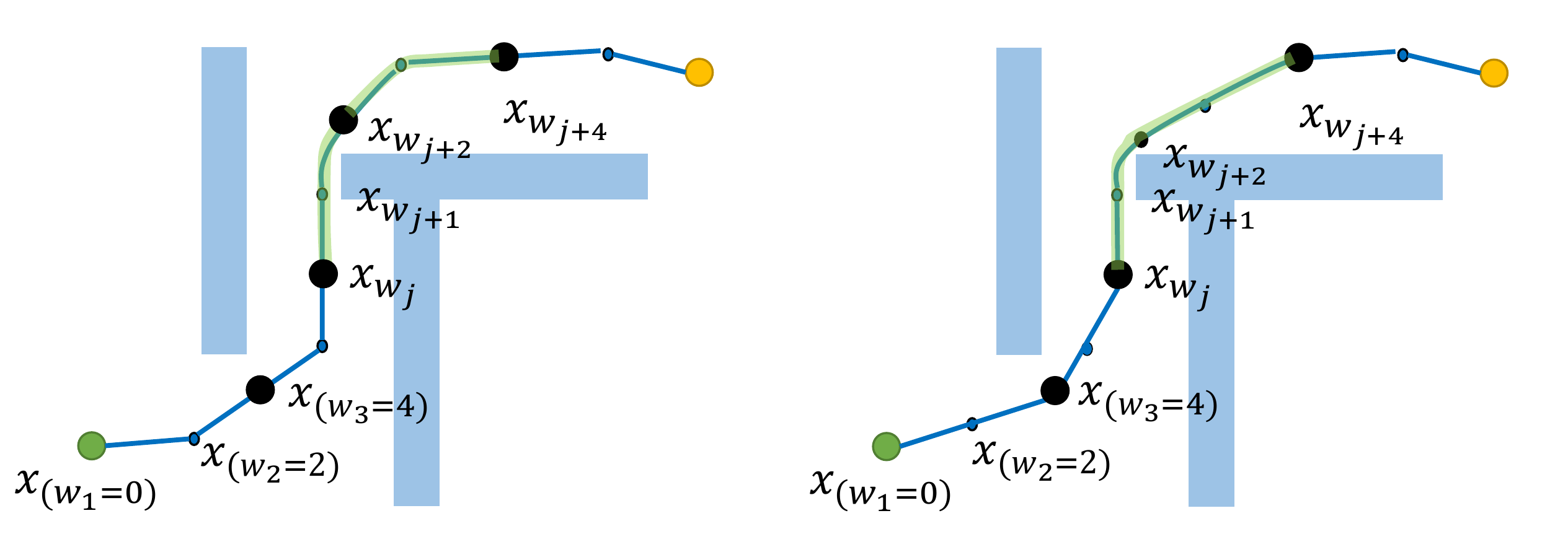}
\caption{Illustration of segment merging.}
\label{fig:merging}
\end{center}
\end{figure}

\noindent\textbf{Trajectory resampling.} Since the path length will be reduced after every iteration, we use $\tt{resample\_traj}$ to find the new planning horizon $H^{(k)}$ according to the desired robot operating speed. We record the new set of split-point indices with $\tt{split\_reference}$.

\noindent\textbf{Termination conditions.} The algorithm terminates if either: (1) the algorithm reaches the maximum number of iterations (i.e., 20 iterations) or (2) the cost of the entire trajectory between iterations is less than the threshold $\epsilon$, i.e., $\|f(\mathbf{x}^{(k)})-f(\mathbf{x}^{(k+1)})\|\leq \epsilon$.

%In the 2D case, the stopping and merging conditions of the algorithm are determined by a chosen constant $\epsilon$, which represents the Euclidean distance between a waypoint $i$ of iteration $k$, $\mathbf{w}_i^{(k)}$, and waypoint $i$ of iteration $k+1$, $\mathbf{w}_i^{(k)}$. (For the manipulator case, the constant represents the difference in joint angles.) After every iteration, check every pair of consecutive segments $(\mathbf{x}_{w_j}^{(k)}, \mathbf{x}_{w_{j+1}}^{(k)})$ in the trajectory $\mathbf{x}^{(k)}$. If $\sum_{i\in ({w_j}^{(k+1)}, w_{j+1}^{(k+1)})} \|x^{(k)}_i - x^{(k+1)}_i\|_2\leq 2\epsilon H   / N_{segments}$, the pair of segments is merged together. Similarly, the algorithm terminates when the average change in distance of a waypoint between iterations is less than $\epsilon$, i.e., $\frac{1}{H}\sum_{i=1}^{H} \|x^{(k-1)}_i - x^{(k)}_i\|_2\leq \epsilon$. (Note: 1-norm for manipulator)

\section{SIMULATION SETUP AND RESULTS}

\subsection{Robot Models}
We use three different robot platforms to test RRT*-sOpt.
\subsubsection{Mobile robot} We model the mobile robot as a point mass on a $2D$-plan. Denote the states of the mobile robot at time step $t$ as $z_{t} = [x_t, y_t]^\top$, the input velocity as $u_{t} = [v_{x,t}, v_{y,t}]^\top$, and the robot configuration as $\theta = [x, y]^\top$. 

\subsubsection{Manipulator} Denote the states of a 5-Dof manipulator as $z_{t} = [\theta_1, \theta_2, \theta_3, \theta_4, \theta_5, \omega_1, \omega_2, \omega_3, \omega_4, \omega_5]_t^\top$, where $\theta_i$ and $\omega_i$, $i \in \left \{ 1,2,3,4,5 \right \}$ are the angle and the angular velocity of the $ith$ joint, respectively. The input contains the angular acceleration at each joint, denoted as $u_{t} = [\alpha_1,\alpha_2,\alpha_3,\alpha_4,\alpha_5]_t^\top$. The robot configuration is $\theta = [\theta_1, \theta_2, \theta_3, \theta_4, \theta_5]^\top$. 

\subsubsection{Mobile manipulator} The kinematic model used for the mobile manipulator combines the model of the arm, which is similar to the manipulator model shown above, and a unicycle model for the mobile base. The state vector at time step $k$ is denoted as $z_t = [x_1,y_1,\theta_1,\theta_2,\theta_3,\theta_4,\theta_5,\omega_2,\omega_3,\omega_4,\omega_5]_k^\top$, where $\theta_1$ is the heading angle of the mobile base, $\theta_i$ and $\omega_i$, $i \in \{2,3,4,5\}$ are the angle and the angular velocity of the arm joints. The input vector is denoted as $u_t = [v, \omega_1, \alpha_2, \alpha_3, \alpha_4, \alpha_5]_k^\top$, which are the robot base velocity, base yaw rate, and angular acceleration of the arm joints. The robot configuration is $\theta = [x_1,y_1,\theta_1,v, \omega_1,\theta_2, \theta_3, \theta_4, \theta_5]^\top$. 

The three robot models can be represent in the non-linear form:
\begin{equation}\label{dy_eq}\small
z_{t+1} = g(z_t,u_t).
\end{equation}

\subsection{The Motion Planning Problem}
In this paper, the goal of the motion planning problem is to plan the command that brings the robot to the goal configuration while avoiding obstacles. We first solve for a path using the parallel RRT* with the configuration $\theta$ defined previously. After getting the path $\pmb{\theta}^{RRT^*}$, an optimization problem for the full trajectory can be formulated. The decision variable for each segment at each time step is the input vector, denoted as $\mathbf{u} := [u_0^\top, u_1^\top,\cdots,u_{H-1}^\top]^\top$, where $H$ is the planning horizon. Similarly, the resulting state vector is $\mathbf{z} := [z_1^\top, z_2^\top ,\cdots,z_{H}^\top]^\top$. Given the initial state, $z_{0}$, we obtain $\mathbf{z}=f_{ki}(z_0,\mathbf{u})$ by concatenating the kinematic function (Eq.~(\ref{dy_eq})) throughout the planning horizon. For simplicity, denote the kinematic function as $f_{ki,z_0}(\mathbf{u}):=f_{ki}(z_0,\mathbf{u})$. In order to obtain the optimal solution $\mathbf{u}^*$ given the constrained feasible set $\Gamma$ and the terminal constraint, the following optimization problem needs to be solved:
\begin{equation}\label{P:MP}
\begin{aligned}
\mathbf{u}^* = \arg\min_{\mathbf{u}}\quad  &f_{z_{0}}(\mathbf{u}),\\
s.t.\quad & f_{ki,z_{0}}(\mathbf{u}) \in\Gamma,\\
\quad & z_{H} = z_{goal}.
\end{aligned}
\end{equation}
The cost function is quadratic and has the form: $f_{z_0}(\mathbf{u}) = \| f_{ki,z_0}(\mathbf{u})-\mathbf{z}_{goal}\|_{Q}^2 + \lambda  \|\mathbf{u}\|_R^2$, which is convex and regular. The first term penalizes the deviation from the goal, and the second term penalizes the input.
\subsection{Implementation of sOpt}
 We denote the segment that starts at split-point $z_{w_{j}}$ as $\mathbf{z}_{[w_{j}, w_{j+2}]}^{(k)}$, the associating input vector as $\mathbf{u}_{[w_{j}, w_{j+2}]}^{(k-1)}$, and the convex feasible set as $\chi_{[w_{j}, w_{j+2}]}^{(k)}$. For simplicity, we drop the subscript in this section. For each segment in each iteration, we update $\mathbf{z}^{(k)} = f_{ki,z_{w_j}}(\mathbf{u}^{(k-1)})$ at iteration $k=2,3\dots$. Notice that $\mathbf{z}^{(1)}$ is determined by $\pmb{\theta}^{RRT^*}$ and $\mathbf{u}^{(0)}$ is initialized as a zero vector. (A more sophisticated way is to initialize $\mathbf{u}^{(0)}$ with a motion generator that commands the robot to track $\pmb{\theta}^{RRT^*}$.) The convex feasible set, $\chi_{[w_{j}, w_{j+2}]}^{(k)}$, is determined by $\mathbf{z}_{[w_{j}, w_{j+2}]}^{(k)}$. Given the feasible set $\Gamma = \bigcap_{\bar{j},t}\{\mathbf{z}: h_{\bar{j}}(z_t) \ge 0\}$ ($\bar{j}$ numerates over obstacles and $t$ numerates over time steps in a segment, i.e., $t = 1,\dots,\bar{H}$), the results of the previous iteration ($\mathbf{u}^{(k-1)}$ and $\mathbf{z}^{(k)}$), and the function $f_{ki,z_{w_{j}}}(\mathbf{u})$, we can construct the convex feasible set as:
\begin{eqnarray}
\chi^{(k)}_{[w_{j}, w_{j+2}]} = \bigcap_{\bar{j},t}\{\mathbf{u}: h'_{\bar{j},t}(\mathbf{u},z^{(k)}_t,\mathbf{u}^{(k-1)}) \ge 0\},
\end{eqnarray} 
where $h'_{\bar{j},t} =h_{\bar{j}}(z^{(k)}_t) + \nabla^{\top} h_{\bar{j}}(z^{(k)}_t)\nabla f_{ki,z_{w_{j}},t}(\mathbf{u}^{(k-1)})(\mathbf{u} - \mathbf{u}^{(k-1)})$. Therefore, the iterative sub-problem for each segment with full notation is as follows:
\begin{equation}\label{P:MP-i}
\begin{aligned}
\mathbf{u}_{[w_{j}, w_{j+2}]}^{*(k)} = \arg\min_{\mathbf{u}}\quad \  &f_{z_{w_{j}}}(\mathbf{u}),\\
s.t.\quad & \mathbf{u} \in\chi^{(k)}(\mathbf{u}_{[w_{j}, w_{j+2}]}^{(k-1)}, \mathbf{z}_{[w_{j}, w_{j+2}]}^{(k)}),\\
\quad & z_{\bar{H}} = z_{w_{j+2}},
\end{aligned}
\end{equation}
where $f_{z_{w_{j}}}(\mathbf{u}) = \| f_{ki,z_{w_{j}}}(\mathbf{u})-\mathbf{z}_{{w_{j+2}}}\|_{Q}^2 + \lambda  \|\mathbf{u}\|_R^2$. 

\subsection{Simulation Setup}
We show the simulation results in the following sections. The simulation is conducted in \texttt{Matlab R2021a} on a desktop with a 3.7GHz Intel Core i9-10900K CPU. Parallel computation can be realized by using the function \texttt{parfor}. The stopping criteria for sOpt are the same. The threshold is set at $\epsilon=H \times 10^{-3}$. Obstacles in these scenarios are either convex or wrapped around with convex geometries.  

\subsection{Simulation Results}
\begin{figure}[t]
\begin{center}
\includegraphics[width=8cm]{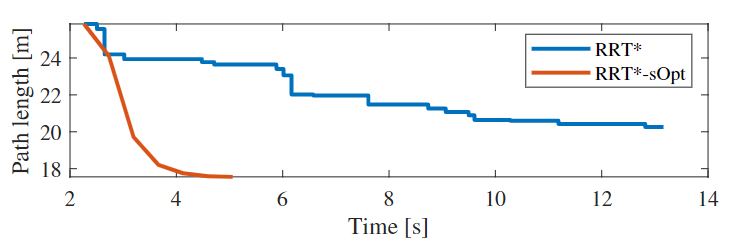}
\caption{Path length reduction performance comparison between RRT* and RRT*-sOpt.}
\label{fig:rrt*}
\end{center}
\end{figure}

\noindent\textbf{Concept verification.} 
One may argue that RRT* alone can also find a near-optimal solution in a short amount of time. Note that RRT* can optimize against different cost. We choose ``path length'' as the cost, which is the simplest cost function in 2D planning. To justify the two stage strategy, we run RRT* for roughly $13$ seconds and compare its path length reduction performance with RRT*-sOpt. As shown in Fig.~\ref{fig:rrt*}, RRT* converges slowly while RRT*-sOpt, using the first returned RRT* solution for initialization, converges quickly to a better local optimal. This confirms the effectiveness of having an optimization solver to improve the solution when a global optimality guarantee isn't necessary. Instead of RRT*, an RRT solution can also serve as an initialization; however, we notice that RRT* can better utilize the samples by rewiring the path without increasing computation time much. Therefore, we adopt RRT* and terminate it once a solution is found. The ``optimization'' of RRT* only happens while RRT* is sampling for the first solution.    

\begin{figure}
 \centering        
   \begin{tabular}{@{}cc@{}}
   \begin{minipage}{.45\linewidth}
    \includegraphics[width=3.5cm]{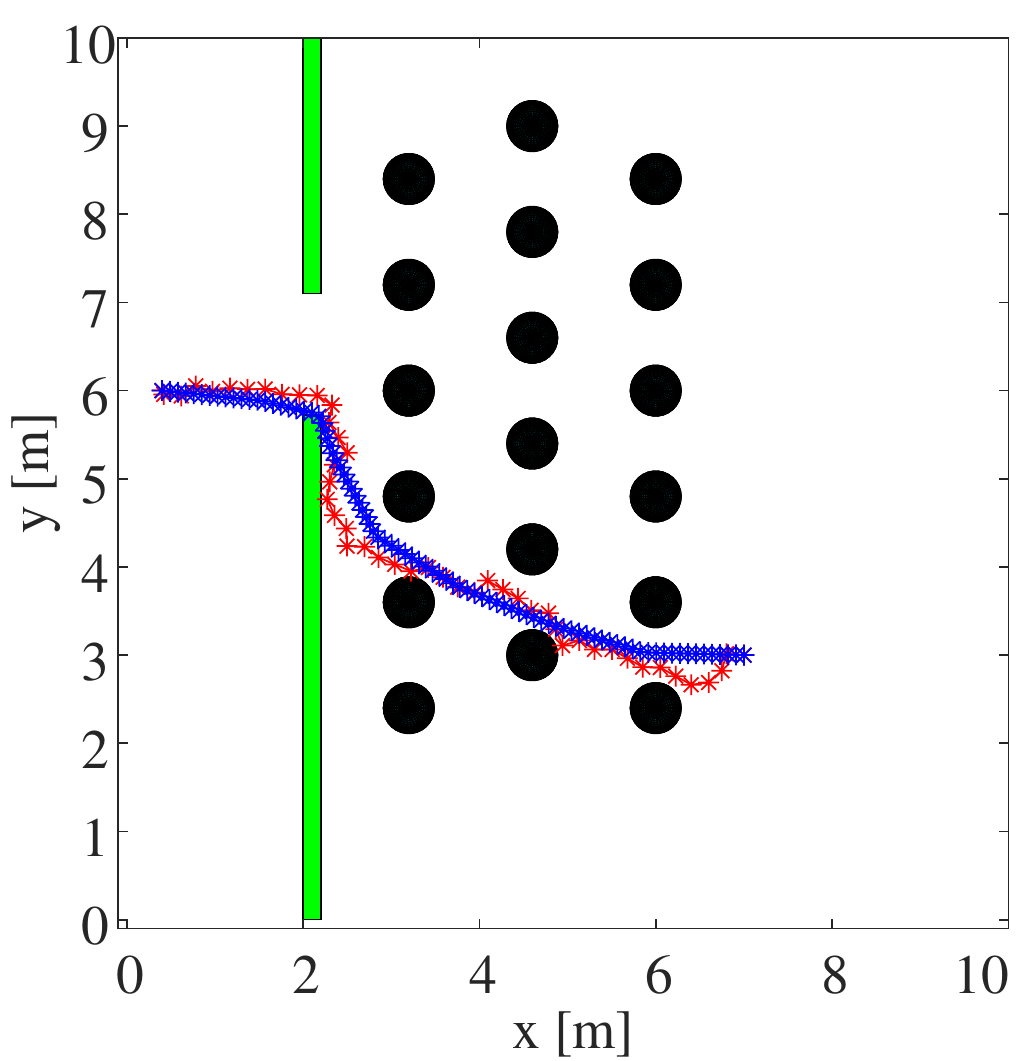}
   \end{minipage} &
    \begin{minipage}{.45\linewidth}
    \includegraphics[width=3.5cm]{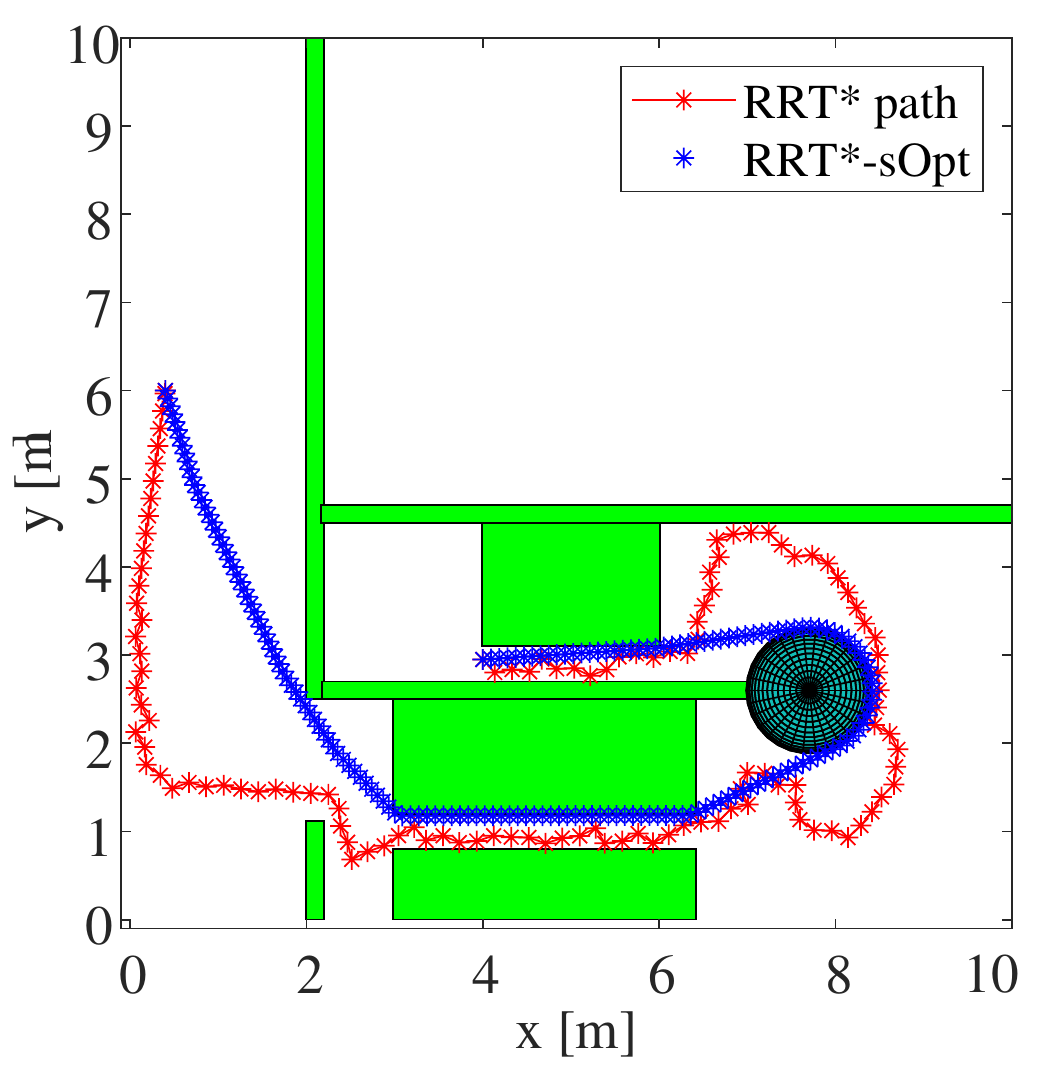}
   \end{minipage}\\ 
   \begin{minipage}{.45\linewidth}
    \includegraphics[width=3.5cm]{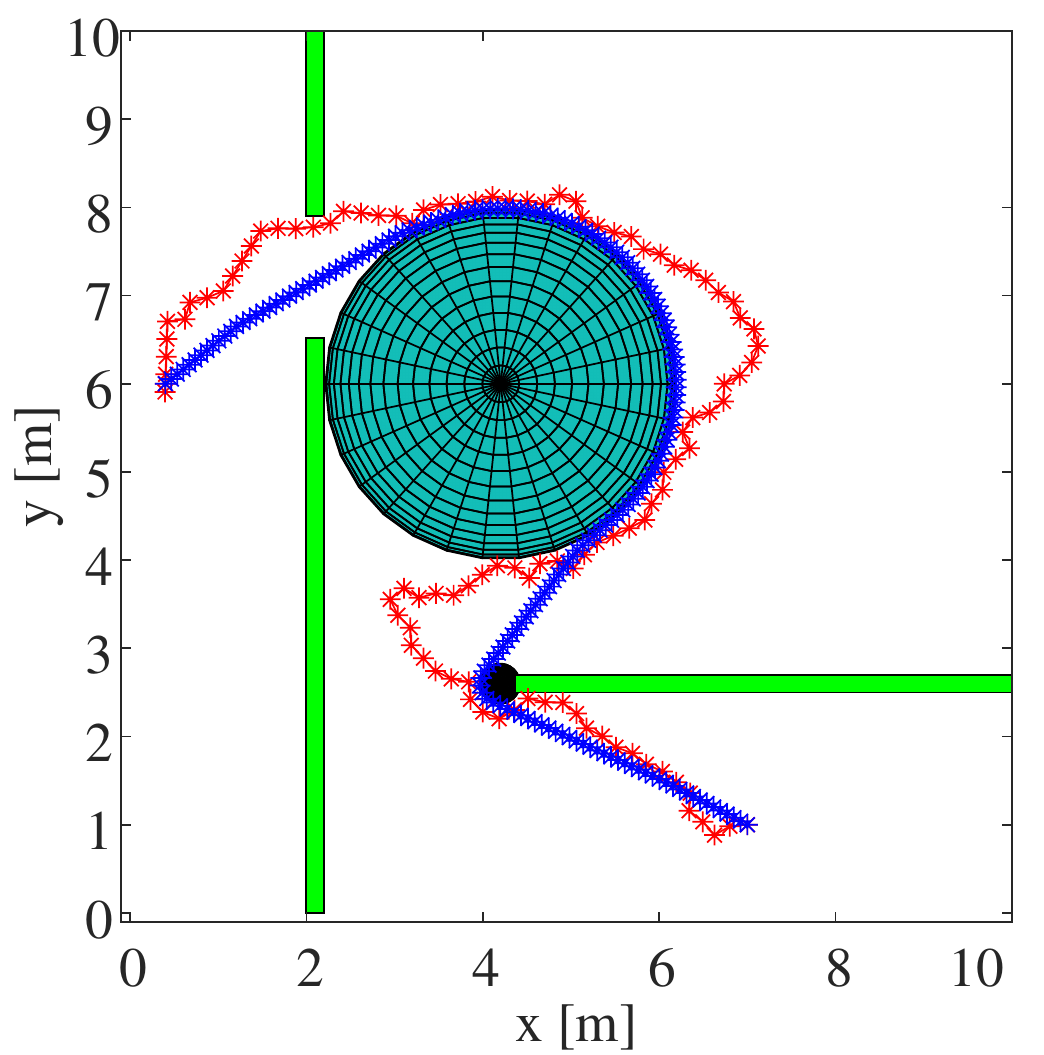}
   \end{minipage} &
    \begin{minipage}{.45\linewidth}
    \includegraphics[width=3.5cm]{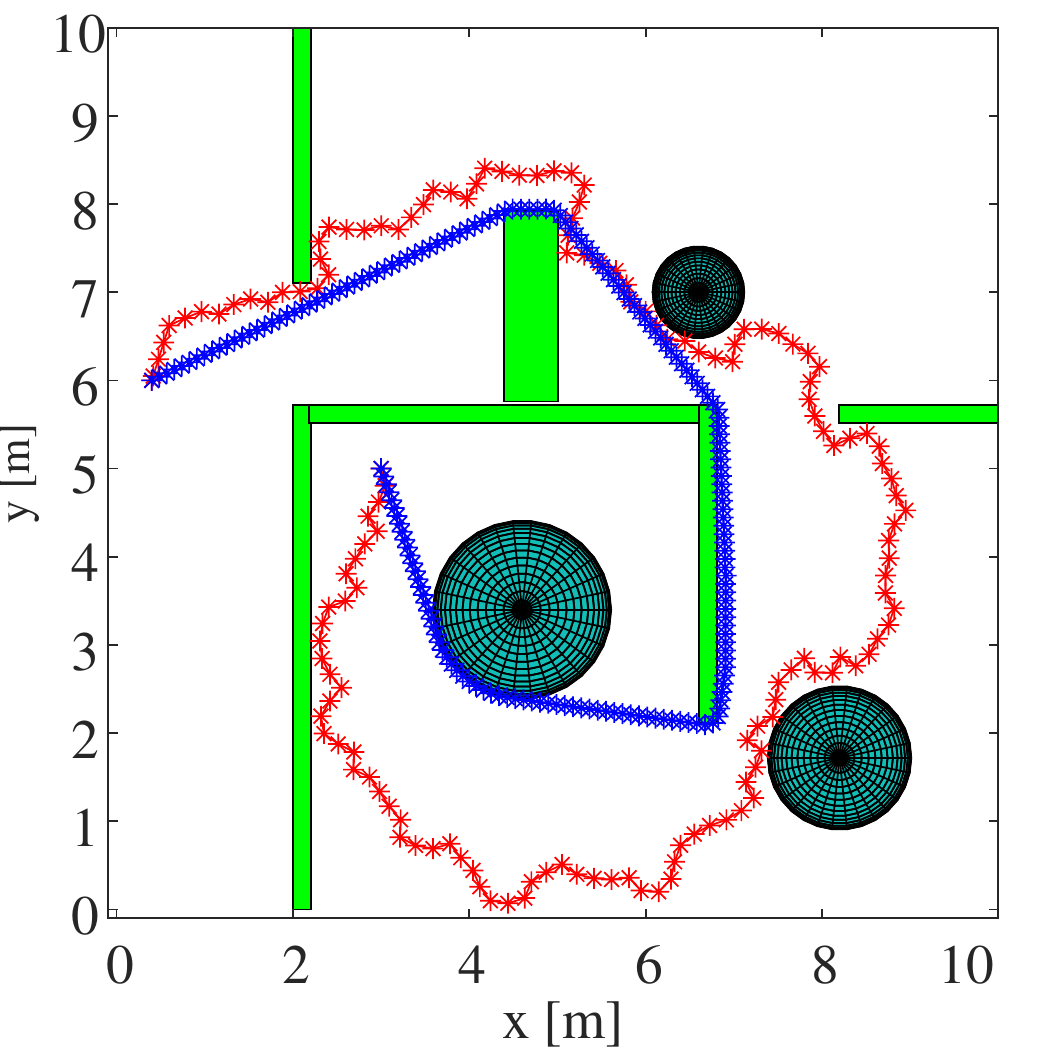}
   \end{minipage}\\ 
  \end{tabular}
        \caption{Simulation results of the 2D motion planning.}
  \label{fig:2D}
\end{figure}

\begin{table*}
\caption{Simulation comparison of 2D planning with $5\sim 20$ obstacles. (The results are the average of 25 trials. The notion ``$7 \rightarrow 3.9$'' in \textbf{\# Segments} in ``auto merge segments'' means that the RRT*-sOpt starting from 7 segments on average terminates at 3.9 segments. The computation time standard deviation for RRT*-sOpt only considered the time variation during the optimization stage.)} \label{table:2D}
\centering
\begin{threeparttable}
\begin{tabular}{L{5cm}|L{1cm}|L{1cm}|L{1cm}|L{1cm}|L{1cm}|L{1cm}|L{1cm}|L{1cm}}
\toprule
\hline
& \multirow{2}{1cm}{RRT*} &\multicolumn{7}{c}{RRT*-sOpt}  \\
\cline{3-9}
& &\multicolumn{5}{c|}{ \# Segments fixed} &\multicolumn{2}{c}{Auto merge segments} \\
\hline
\textbf{\# Segments} &  & $1$ & $3$ & $5$ & $7$ & $9$ & $7\rightarrow 3.9$& $9\rightarrow 5.8$\\ \hline
    \textbf{Computation time average [s]} & $19.82$ & $80.32$ & $22.32$ & $21.63$ &   $21.43$ & $22.01$ & $\mathbf{21.42}$ & $21.81$ \\
    \hline
    \textbf{Computation time standard deviation [s]} &  & $36.93$ & $1.39$ & $1.01$ &   $0.98$ & $1.18$ & $0.93$ & $1.03$ \\
    \hline
    \textbf{Cost} &$37.69$  & $31.24$ & $31.10$ & $30.82$ & $31.08$ & $30.93$ & $31.10$ & $31.16$\\
    \hline
    \textbf{\# iterations} &  & $13.8$ & $6.5$ & $5.9$ & $5.5$ & $7.7$ & $5.5$ & $7.1$\\
    \hline
    \textbf{Success rate ($\%$)} & $100$ & $100$ & $100$ & $100$ & $100$ & $100$ & $100$ & $100$\\
\hline
\bottomrule
\end{tabular}
\end{threeparttable}
\vspace{-15pt}
\end{table*}

\noindent\textbf{2D motion planning.} The scenarios here are designed to simulate long-distance 2D motion planning scenes of mobile platforms. Some of the planning results are shown in Fig.~\ref{fig:2D}. Note that most of our test cases require the robot to travel a long distance, i.e., more than 100 waypoints is needed for a motion plan. The performance comparison of RRT*-sOpt with different numbers of segments is shown in TABLE~\ref{table:2D}. (The computation time for RRT*-sOpt includes both the time spent during the RRT* stage and the optimization stage. The sOpt time can be obtained by subtracting the RRT* time from the total time.) First, we observe that all RRT*-sOpt are able to smooth the RRT* reference and achieve similar final costs, which are noticeably smaller than the costs of the original RRT* solutions. This empirically shows that the algorithm can converge to a local optimum given the RRT* reference. Second, all of the RRT*-sOpt are much quicker than RRT*-Opt. This verifies our claim that trajectory splitting indeed reduces computation time by leveraging parallel computation power. Without merging, $s=7$ has the smallest computation time and number of iterations. On the other hand, the computation time is further reduced when merging is allowed. (The notion $7 \rightarrow 3.9$ means that the RRT*-sOpt starting from 7 segments on average terminates at 3.9 segments.)

\begin{figure*}
 \centering        
      \begin{tabular}{@{}ccc@{}}
   \begin{minipage}{.32\textwidth}
    \includegraphics[width=5.5cm]{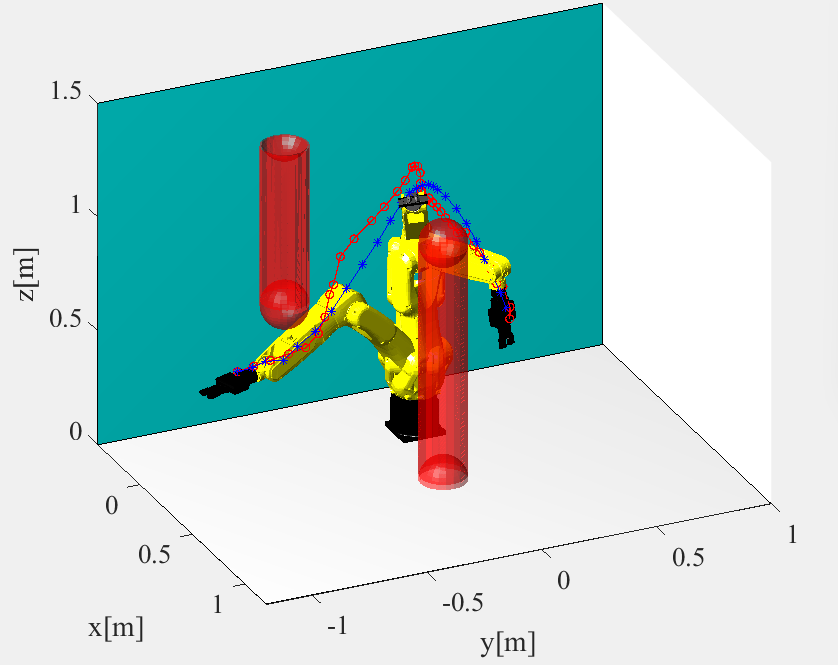}
   \end{minipage} &
    \begin{minipage}{.32\textwidth}
    \includegraphics[width=5.5cm]{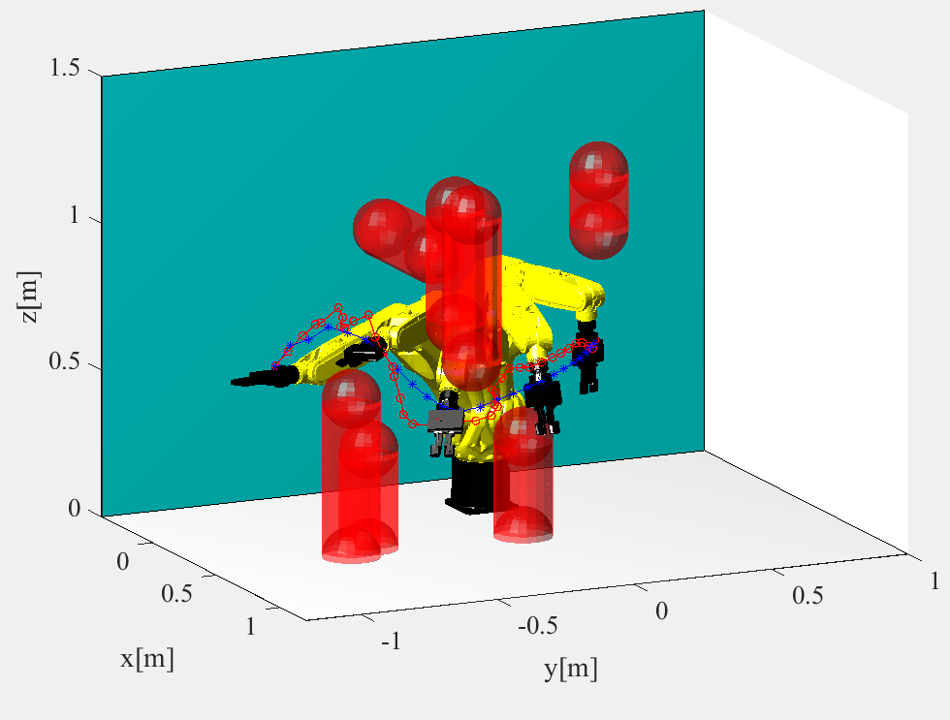}
   \end{minipage}&
    \begin{minipage}{.32\textwidth}
    \includegraphics[width=5.5cm]{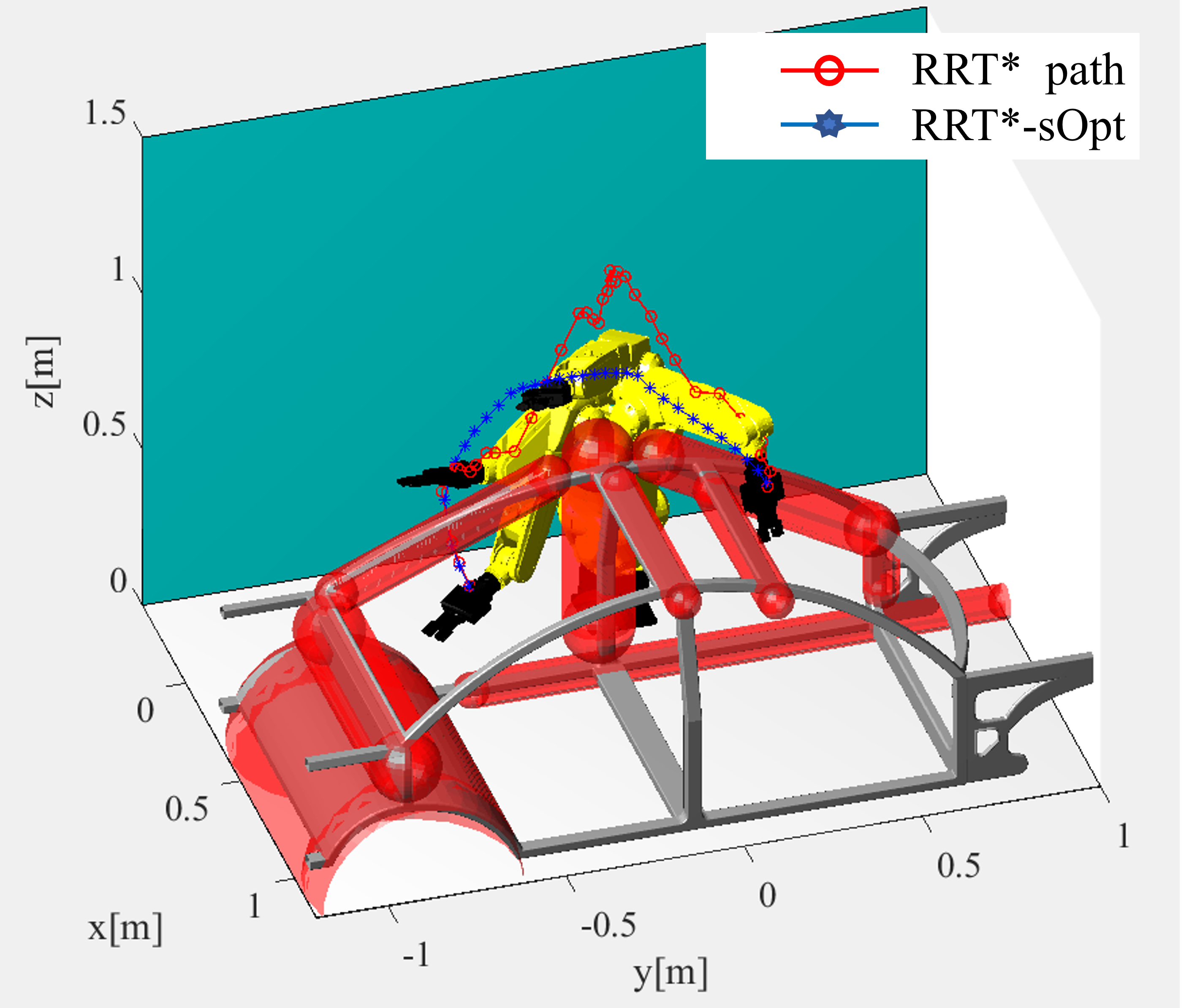}
   \end{minipage}
  \end{tabular}
        \caption{Simulation results of the 5-Dof manipulator motion planning.}
  \label{fig:5D}
\end{figure*}

\begin{table*}
\caption{Simulation comparison of 5-Dof manipulator motion planning. (Average of 20 trials.)}\label{table:5D}
\centering
\begin{threeparttable}
\begin{tabular}{L{5cm}|L{1cm}|L{1cm}|L{1cm}|L{1cm}|L{1cm}|L{1cm}|L{1cm}|L{1cm}|L{1cm}}
\toprule
\hline
\multirow{2}{0.8cm}{} & \multicolumn{9}{c}{\# obstacles: $2$} \\
\cline{2-10}
& \multirow{2}{1cm}{RRT*} &\multicolumn{8}{c}{RRT*-sOpt}  \\
\cline{3-10}
& &\multicolumn{5}{c|}{ \# Segments fixed} &\multicolumn{3}{c}{Auto merge segments} \\
\hline
\textbf{\# Segments} &  & $1$ & $3$ & $5$ & $7$ & $9$ & $5 \rightarrow 3.6$& $7 \rightarrow 6.2$ & $9 \rightarrow 8.4$\\ \hline
    \textbf{Computation time average [s]} & $1.49$   & $98.87$ & $4.66$ & $4.16$ &   $3.80$ & $3.65$ & $4.14$ & $3.78$ & $\mathbf{3.57}$ \\
    \hline
    \textbf{Computation time standard deviation [s]} &   & $84.23$ & $1.29$ & $0.90$ &  $0.93$ & $0.76$ & $1.06$ & $0.76$ & $1.10$ \\
    \hline
    \textbf{Cost} &$8.40$  & $10.4$ & $7.76$ & $7.76$ & $7.73$ & $7.77$ & $7.75$ & $7.74$ & $7.77$ \\
    \hline
    \textbf{\# iterations} &  & $6.2$ & $5.6$ & $6.4$ & $7.2$ & $6$ & $6.4$ & $6.8$ & $6$ \\
    \hline
    \textbf{Success rate ($\%$)} & $100$ & $60$ & $100$ & $100$ & $100$ & $100$ & $100$ & $100$ & $100$\\
\hline
\multirow{2}{0.8cm}{} & \multicolumn{9}{c}{\# obstacles: $7 \sim 9$} \\
\cline{2-10} 
& \multirow{2}{1cm}{RRT*} &\multicolumn{8}{c}{RRT*-sOpt}  \\
\cline{3-10}
& &\multicolumn{5}{c|}{ \# Segments fixed} &\multicolumn{3}{c}{Auto merge segments} \\
\hline
\textbf{\# Segments} &  & $1$ & $3$ & $5$ & $7$ & $9$ & $5 \rightarrow 3$& $7 \rightarrow 4.7$ & $9 \rightarrow 6.3$\\ \hline
    \textbf{Computation time average [s]} & $23.26$ & $170.61$ & $28.42$ & $27.48$ &   $26.56$ & $26.11$ & $25.77$ & $\mathbf{25.76}$ & $25.79$ \\
    \hline
    \textbf{Computation time standard deviation [s]} &   & $48.51$ & $2.23$ & $1.04$ &   $0.56$ & $0.12$ & $0.59$ & $0.10$ & $1.00$ \\
    \hline
    \textbf{Cost} &$9.35$  & $7.80$ & $7.86$ & $7.57$ & $7.58$ & $7.53$ & $7.69$ & $7.66$ & $7.63$ \\
    \hline
    \textbf{\# iterations} &  & $6.4$ & $6$ & $7.9$ & $7.3$ & $7.6$ & $6.3$ & $7$ & $7.2$ \\
    \hline
    \textbf{Success rate ($\%$)} & $100$ & $10$ & $100$ & $100$ & $100$ & $100$ & $100$ & $100$ & $100$\\
\hline
\bottomrule
\end{tabular}
\end{threeparttable}
\end{table*}

\noindent\textbf{Motion planning for a 5-Dof manipulator.} The scenarios here are designed to simulate general 3D motion planning scenes of manipulators operating in factories. Two categories of settings are created: one with only two obstacles; the other with $7 \sim 9$ obstacles. Some of the planning results are shown in Fig.~\ref{fig:5D}. The performance comparison is shown in TABLE~\ref{table:5D}. Similar to the 2D case, RRT*-sOpt is much quicker than RRT*-Opt. Note that RRT*-Opt fails in some of the test cases with two obstacles and fails in most of the cases with more obstacles. This is mainly due to the linearization and accumulating approximation errors when solving optimization problems. Without segmenting the trajectory, the optimization solver will need to handle all the constraints at once, resulting in a larger chance of failure. On the other hand, RRT*-sOpt mitigates the problem caused by accumulating linearization errors by distributing the environment (safety) constraints to multiple segments. Also, the computation time of RRT*-sOpt is less sensitive to the number of obstacles compared to that of RRT*-Opt, i.e., RRT*-sOpt performs more robustly against different configurations of the environment. The reduction in computation time due to merging is more evident in manipulator motion planning problems, especially in the second category of settings. In the cases with $s$ going from $7 \rightarrow 4.7$ (on average), the computation speed is $12 \%$ faster than the fastest RRT*-sOpt without merging. %We also tested the case with $s=11$. While the computation time future reduced to $1.60$, it tends to terminate at solutions with higher costs without merging. 

\begin{table*}
\caption{Simulation comparison of mobile manipulator motion planning. (Average of 10 trials.)}\label{table:MM}
\centering
\begin{threeparttable}
\begin{tabular}{L{5cm}|L{1cm}|L{1cm}|L{1cm}|L{1cm}|L{1cm}|L{1cm}}
\toprule
\hline
& \multirow{2}{1cm}{RRT*} &\multicolumn{5}{c}{RRT*-sOpt}  \\
\cline{3-7}
& &\multicolumn{4}{c|}{ \# Segments fixed} &\multicolumn{1}{c}{Auto merge} \\
\hline
\textbf{\# Segments} &  & $3$ & $5$ & $7$ & $9$ & $7\rightarrow 4$\\ \hline
    \textbf{Computation time average [s]} & $16.45$ & $32.32$ & $22.64$ & $\mathbf{21.67}$ &   $21.84$ & $24.16$ \\
    \hline
    \textbf{Computation time standard deviation [s]} &   & $3.73$ & $2.01$ & $1.71$ &   $1.99$ & $3.37$ \\
    \hline
    \textbf{Cost} &$39.73$  & $34.91$ & $22.27$ & $22.65$ & $23.14$ & $22.46$ \\
    \hline
    \textbf{\# iterations} &  & $9$ & $7.8$ & $8$ & $9.2$ & $7.6$ \\
    \hline
    \textbf{Success rate ($\%$)} & $100$ & $60$ & $100$ & $100$ & $100$ & $100$\\
\hline
\bottomrule
\end{tabular}
\end{threeparttable}
\vspace{-15pt}
\end{table*}

\begin{figure}[t]
\begin{center}
\includegraphics[width=8cm]{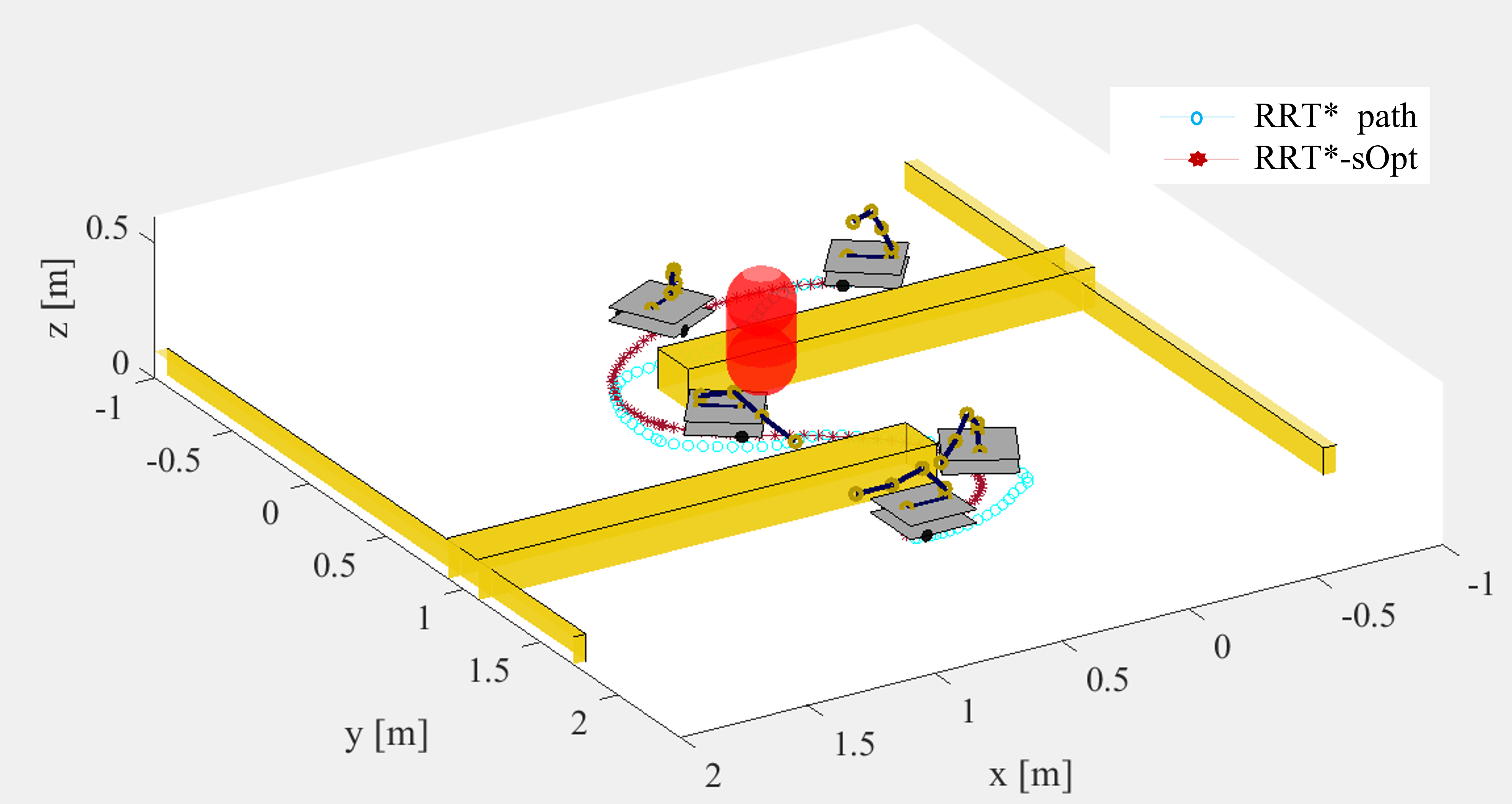}
\caption{Simulation results of a mobile manipulator motion planning.}
\label{fig:MM}
\end{center}
\end{figure}

\noindent\textbf{Motion planning for a mobile manipulator.} The scenarios here are designed to simulate 3D motion planning scenes of mobile manipulators traveling through hallways while moving the arm to avoid obstacles. One of the planning results is shown in Fig.~\ref{fig:MM}. The performance comparison is shown in TABLE~\ref{table:MM}. Since the mobile manipulator kinematic model is more complicated, we formulate the non-linear constraints and use an open-source solver, CasADi \cite{Andersson2019}, to solve the nonlinear planning problem directly. Nevertheless, RRT*-Opt still fails in most cases because CasADi also requires approximations during the solving process, and the solver is still likely to fail when too many constraints are included in one optimization problem. On the other hand, with segmentation, this problem can be mitigated and the computation advantage increases as the number of segment increases, similar to what we observe in the 2D and manipulator cases. This also indicates that sOpt can reduce computation time when working with other solvers. Notice that, in this case, RRT*-sOpt with merging does not reduce the computation time. This is mainly due to the overhead required to setup CasADi when merging occurs. In the future, we will improve the implementation (by choosing a different optimization solver or coding language) to verify the performance of RRT*-sOpt with merging.

In summary, simulation results show that RRT*-sOpt can successfully plan long-horizon motion plans for mobile robots, manipulators, and mobile manipulators. With the trajectory segmentation, the computation time is significantly reduced and is relatively robust to different number of obstacles. The novel idea of segment merging is also tested in these settings and has demonstrated potential to further reduce the computation time. Based on the results with the three models, we suggest using auto-merge-segment RRT*-sOpt for 2D and manipulator planning, and fixed-segment RRT*-sOpt for mobile manipulator planning. (The number of segments can be determined based on the planning horizon. Empirically, we suggest that each segment should not handle more than 30 time steps.) The computation time standard deviation of RRT*-sOpt is also small compared to RRT*-Opt. It is worth noticing that the setups that have a small average time also tend to have a smaller standard deviation.

\subsection{Discussion and future work}
Though the current implementation of RRT*-sOpt has improved the computation time substantially, there are still some areas that remain to be improved. 
\begin{itemize}
  \item A method of selecting the initial number of segments is required. One way of determining such a number is to choose a horizon for the initial segments. Then, by calculating the path length of the RRT* reference, we can determine the number of segments needed. However, this selection method does not consider the configuration of the environment (e.g., obstacles' relative locations). In the future, we aim to develop a method that determines the initial number of segments based on both reference path length and environment configuration.  
  \item The merging condition can be improved. The current implementation uses the cost reduction trend to determine when the merging occurs. However, we observed that the optimization problem with tight space usually converges in fewer iterations. This indicates that the environment configuration should be directly taken into account when designing the merging conditions.  
  \item Although the algorithm converges empirically, we hope to investigate the theoretical properties of RRT*-sOpt. 
\end{itemize}

\section{CONCLUSION}
This paper presented a fast long-horizon motion planning algorithm, RRT*-sOpt, that inherits the computation advantages of its predecessor, RRT*-CFS, and further improves it by incorporating the idea of segmented trajectory optimization. The RRT*-sOpt quickly found a feasible and semi-optimal path using RRT* and iteratively refined the solution using sOpt. Simulation results showed that RRT*-sOpt benefits from the hybrid structure and the ability to distribute the problem complexity to leverage the power of parallel computation. RRT*-sOpt can solve problems that are extremely challenging to stand-alone optimization-based planners, has better final cost compared to pure sampling-based planners, and has significantly shorter computation time compared to previous hybrid planners. The novel idea of segment merging was also tested and has shown potential to further reduce the computation time. We conclude that the hybrid structure with trajectory segmentation has indeed brought strong performance to RRT*-sOpt for general long-horizon robot motion planning problems. 

\section*{Acknowledgement}
The authors thank Changliu Liu for helpful discussions. This work was supported by the National Science Foundation under Grant No.1734109. Any opinion, finding, and conclusion expressed in this paper are those of the authors and do not necessarily reflect those of the National Science Foundation. 

\FloatBarrier
\bibliography{root(ECC2021)}
\bibliographystyle{IEEEtran}

\end{document}